\documentclass{llncs}
\usepackage{graphicx}
\begin{document}

\textbf{This is a preprint. Please, cite us as:}
\\
A.L. Alfeo, M.G.C.A. Cimino, S. Egidi, B. Lepri, A. Pentland, G. Vaglini, "Stigmergy-Based Modeling to Discover Urban Activity Patterns from Positioning Data" in Proc. SBP-BRiMS The International Conference on Social Computing, Behavioral-Cultural Modeling and Prediction and Behavior Representation in Modeling and Simulation (SBP-BRiMS 2017), LNCS 10354, pp. 292-301, Washington, DC, USA, 5-8 July, 2017, (ISBN 978-3-319-60239-4) $DOI: 10.1007/978-3-319-60240-0\_35$

\newpage

\title{Stigmergy-based modeling to discover urban activity patterns from positioning data.}
\titlerunning{Stigmergy-basedModeling}  					  
%
\author{Antonio L. Alfeo\inst{1} \and Mario G. C. A. Cimino\inst{1}
Sara Egidi\inst{1} \and Bruno Lepri\inst{2} \and Alex Pentland\inst{3} \and Gigliola Vaglini\inst{1}}
\authorrunning{Antonio L. Alfeo et al.} 
%
\tocauthor{Antonio L. Alfeo, Mario G. C. A. Cimino, Sara Egidi, Bruno Lepri, Alex Pentland, Gigliola Vaglini}
\institute{
University of Pisa, largo Lazzarino 1, Pisa, Italy\\
\email{luca.alfeo@ing.unipi.it, mario.cimino@unipi.it, s.egidi1@studenti.unipi.it, gigliola.vaglini@unipi.it}\\ 
\and
Bruno Kessler Foundation, via S. Croce, 77, Trento, Italy\\
\email{lepri@fbk.eu}\\
\and
M.I.T. Media Laboratory, Cambridge, MA 02142, USA\\
\email{pentland@media.mit.edu}
}

\maketitle              

\begin{abstract}
Positioning data offer a remarkable source of information to analyze crowds urban dynamics. However, discovering urban activity patterns from the emergent behavior of crowds involves complex system modeling. An alternative approach is to adopt computational techniques belonging to the emergent paradigm, which enables self-organization of data and allows adaptive analysis. Specifically, our approach is based on stigmergy. By using stigmergy each sample position is associated with a digital pheromone deposit, which progressively evaporates and aggregates with other deposits according to their spatiotemporal proximity. Based on this principle, we exploit positioning data to identify high-density areas (hotspots) and characterize their activity over time. This characterization allows the comparison of dynamics occurring in different days, providing a similarity measure exploitable by clustering techniques. Thus, we cluster days according to their activity behavior, discovering unexpected urban activity patterns. As a case study, we analyze taxi traces in New York City during 2015.

\end{abstract}
\section{Introduction}
The increasing volume of urban human mobility data arises unprecedented opportunities to monitor and understand crowd dynamics. Identifying events which do not conform to the expected patterns can enhance the awareness of decision makers for a variety of purposes, such as the management of social events or extreme weather situations \cite{sagl:2012}. 
For this purpose GPS-equipped vehicles provide a huge amount of reliable data about urban human mobility, exhibiting correlation with people daily life, events, and city structure \cite{veloso:2011}. 
The majority of the methods approaching the analysis of vehicle traces can be grouped into three categories: \emph{cluster-based}, \emph{classification-based}, and \emph{pattern mining-based}; whereas the main application problems include the hotspot discovery, the extraction of mobility profiles, and the detection and monitoring of big events and crowd behavior \cite{mazimpaka:2016}.
For example, in \cite{zhang:2015} the impact of a social event is evaluated by analyzing taxi traces. Here, the authors model typical passenger flow in an area, in order to compute the probability that an event happens. Then, the event impact is measured by analyzing abnormal flows in the area via Discrete Fourier Transform. In \cite{pan:2013} GPS trajectories are mapped through an Interactive Voting-based Map Matching Algorithm. This mapping is used for off-line characterization of normal drivers' behavior and real-time anomaly detection. Furthermore, the cause of the anomaly is found exploiting social network data. In \cite{kuang:2015} the authors use a Multiscale Principal Component Analysis to analyze taxi GPS data in order to detect traffic congestion.  

One of the main issues concerning the analysis of this kind of data is their dimensionality. Many approaches handle it by focusing on specific areas (\emph{hotspots}) whose high concentration of events and people can summarize mobility dynamics \cite{hu:2014}. As an example, in \cite{lu:2016} a density-based spatial clustering is employed to perform spatiotemporal analysis on taxi pick-up/drop-off to find seasonal hotspots. Authors in \cite{keler:2016} use OPTICS algorithm in order to detect city hotspots as density-based clusters of taxi drop-off positions. Recently, in \cite{li:2012} an Improved Auto-Regressive Integrated Moving Average algorithm is proposed; it is aimed to detect urban mobility hotspots via taxi GPS traces and analyze the dynamics of pick-ups in dense locations of the city. 
However, due to the complexity of human mobility data, the modeling and comparison of their dynamics over time remain hard to manage and parametrize \cite{castro:2013}. 
In this paper, we present an innovative approach based on \emph{stigmergy} \cite{marsh:2008} that aims to handle both complexity and dimensionality of these data, providing an analysis of urban crowds dynamics by exploiting taxi GPS data. Specifically, our investigation covers the city hotspots identification, the characterization of their activity over time and the unfolding of unexpected activity pattern.   

The paper is structured as follows. In Section 2 the architectural view of our approach is described. In Section 3 the experimental studies and results are presented. Finally, Section 4 summarizes conclusions and future work.

\section{Approach  Description}
In this section, we present our approach, based on the principle of \emph{stigmergy}. Stigmergy is an indirect coordination mechanism used in social insect colonies \cite{marsh:2008}. It is based on the release of chemical markers (\emph{pheromones}), which aggregate when subsequently deposited in proximity with each other. This mechanism can be employed in the context of data processing, providing self-organization of data \cite{vernon:2007} while unfolding their spatial and temporal dynamics \cite{barsocchi:2015}. By exploiting stigmergy, we discover city hotspots, characterize their activity dynamics (i.e. presence of people over time) and assess unexpected activity patterns. In order to focus on activity dynamics, we employ New York City taxi positioning data, considering the amount of passengers together with the GPS position of each pick-up/drop-off.

\subsection{Hotspot Detection}
\begin{figure}
  \centering
  \includegraphics[width=\textwidth]{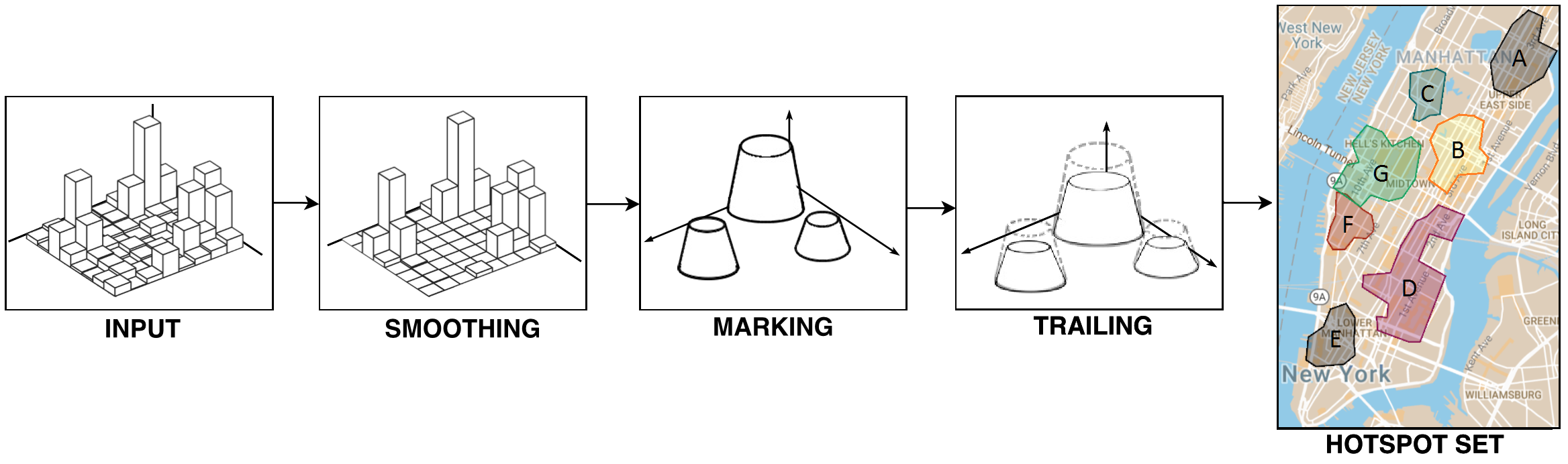}
  \caption{The stigmergy-based process of hotspot discovery.}
  \label{img:HotspotDiscovery}
\end{figure}

At the beginning, data samples are transformed in digital pheromone deposits, allowing the progressive emergence of city hotspots (i.e. the most high-density areas within the city). Firstly, data are treated by the smoothing process (Fig.\ref{img:HotspotDiscovery}), in order to remove insignificant activity levels and highlight relevant dynamics. This process is implemented by applying a sigmoidal function to the samples. 
Then, a mark is released in correspondence of each smoothed sample in a three-dimensional virtual environment. Marks are defined by a truncated cone with a given width and intensity (height) equal to data sample value. The trailing process aggregates marks, forming a \emph{stigmergic trail}, which is characterized by evaporation (i.e. temporal decay $\delta$). Eq. \ref{eq:trail} describes the trail at time instant $i$.

\begin{equation}
\label{eq:trail}
T_{i} = (T_{i-1} - \delta) + Mark_{i}
\end{equation}

As an effect, isolated marks tend to disappear, whereas the arrival of new marks in a given region counteracts the evaporation. Thus, aggregation and evaporation can act as an agglomerative spatiotemporal clustering with historical memory.
Hotspots are identified as the city areas corresponding to the overlapping of the most relevant trails obtained by processing data in early morning (i.e. 3am-8am), morning (i.e. 9am-2pm), afternoon/evening (i.e. 3pm-8pm), and night (i.e. 9pm-2am) time slots. As an example, Fig. \ref{img:HotspotDiscovery} shows the hotspots identified in Manhattan (New York City). Their locations correspond to: East Harlem - Upper East Side (A), Midtown East (B), Broadway (C), East Village - Gramercy - MurrayHill (D), Soho - Tribeca (E), Chelsea (F) and Time Square - Midtown West - Garment (G).

\subsection{Hotspot Activity Characterization}
For each identified hotspot, we generate the activity time series, by periodically collecting the amount of activity occurred in the hotspot during a day. Let us consider an activity time series; what is actually interesting is not the continuous variation of the activity over time, but the transition from one type of behavior to another. 

Generally, given a time window each hotspot behavior can be characterized by an ideal time series segment of hotspot activity representing that specific behavior. More formally, we define it as an \emph{archetype}. An example of an archetype is \emph{asleep} behavior, which usually occurs during the night, between the calming down of the nightlife and the arrival of the workers; here the city exhibits its lowest activity level.

\begin{figure}
  \centering
  \includegraphics[width=0.77\textwidth]{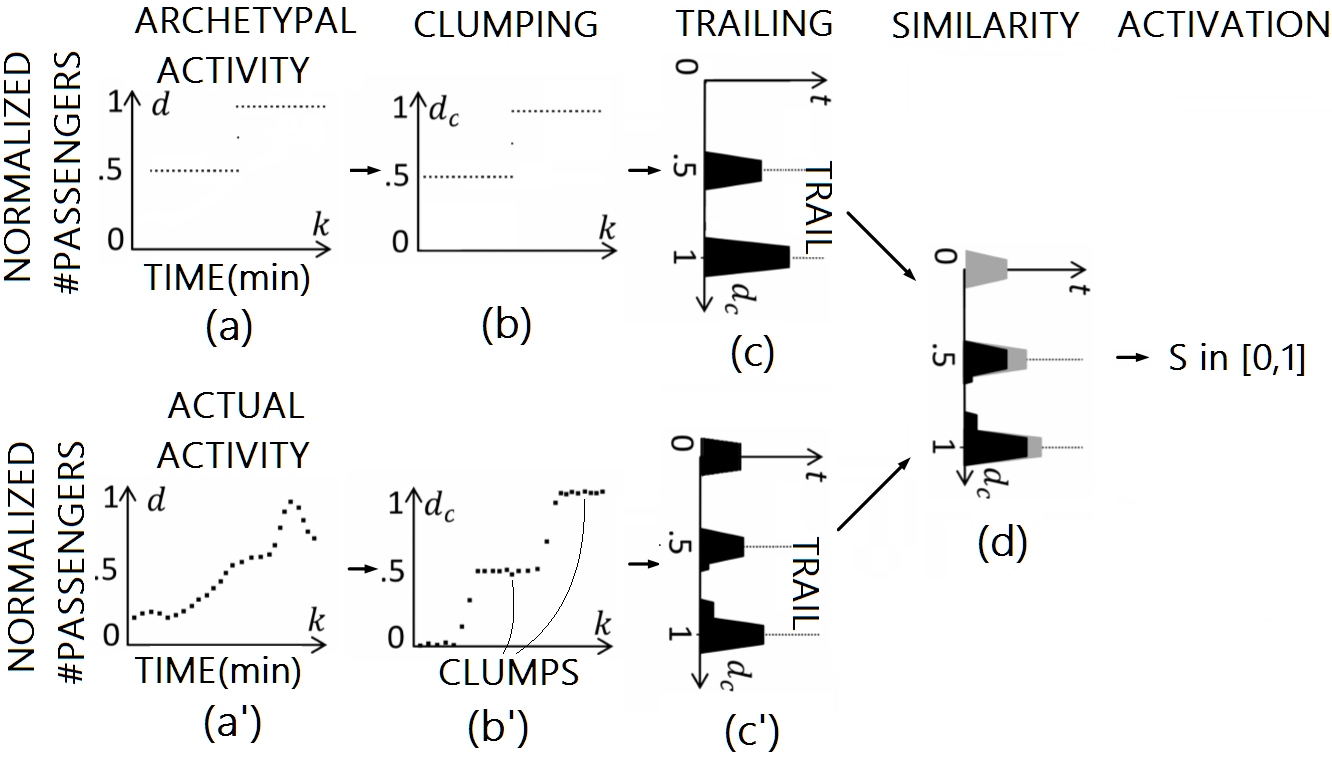}
  \caption{The architecture of a SRF.}
  \label{img:srf}
\end{figure}

In order to detect an archetypal behavior in hotspot activity time series, we design a processing schema called Stigmergic Receptive Field (SRF), because it is receptive to a specific archetype and it processes samples employing the principle of stigmergy. Specifically, SRF computes a degree of similarity between a specific archetype (Fig. \ref{img:srf}a) and an activity time series (Fig. \ref{img:srf}a'), by subsequently processing their samples, which are assumed to be normalized between 0 and 1. 

First, samples undergo the \emph{clumping process} (Fig. \ref{img:srf}b and Fig. \ref{img:srf}b'), which acts as a sort of soft discretization creating clumps of samples. Clumps arrangement can be parametrized  allowing to fit the analysis over the archetype's levels of interest. The clumping can be implemented as a double sigmoidal function. Second, the marking process (Fig. \ref{img:srf}c and Fig. \ref{img:srf}c') enables the release of a mark in a bi-dimensional virtual environment in correspondence of the sample value. The mark can be implemented by a trapezoid with given intensity (height) and width $\epsilon$. Third, the trailing process accumulates marks creating the trail structure, whose intensity decays (i.e. evaporates) of a given rate $\delta$ at each step of time.
As an effect, evaporation rate and mark width allow the trail to capture coarse spatiotemporal structure in data, handling micro-fluctuations. Fourth, current $T_{act}$ 
and archetypal trails $T_{arc}$ are compared by the similarity process (Fig. \ref{img:srf}d), by using the Jaccard coefficient \cite{jaccard:2013} as defined in Eq.  \ref{eq:jaccard}.

\begin{equation}
\label{eq:jaccard}
S = \frac{|T_{arc} \cap T_{act}|}{|T_{arc} \cup T_{act}|}
\end{equation}

This coefficient provides a measure of similarity between 1 (identical trails) and 0 (non-overlapping trails). 
Finally, the activation process is applied to enhance only relevant similarity values and remove insignificant values according to the activation thresholds $\alpha_{a},\beta_{a}$. This process can be implemented by using the already mentioned sigmoidal function (Eq. \ref{eq:sigm}). 

\begin{equation}
\label{eq:sigm}
f(x,\alpha_{a},\beta_{a}) = \frac{1}{(1+e^{-\alpha_{a}(x-\beta_{a})})}
\end{equation}

In order to provide an effective similarity, the SRF's parameters have to be properly tuned. With this aim, the Adaptation process uses the Differential Evolution (DE) to adapt the structural parameters of the SRF: (i) the clumping inflection points $\alpha$, $\beta$, $\gamma$, $\lambda$; (ii) the mark width $\epsilon$; (iii) the trail evaporation $\delta$; (iv) the activation thresholds $\alpha_a$, $\beta_a$. The aim of DE is to minimize the mean square error (MSE), considering the error as the difference between the target $\hat S$ 
and the computed $S$ similarity values over a set of M labeled time series (Eq. \ref{eq:mse}).

\begin{equation}
\label{eq:mse}
Fitness = \frac{\sum_{i=1}^{M} (|S _{i} -\hat S _{i}|^2)}{M}
\end{equation}

The target similarity value is 1 if the current time series exhibits the archetypal behaviour, 0 otherwise.

\begin{figure}
  \centering
  \includegraphics[width=\textwidth]{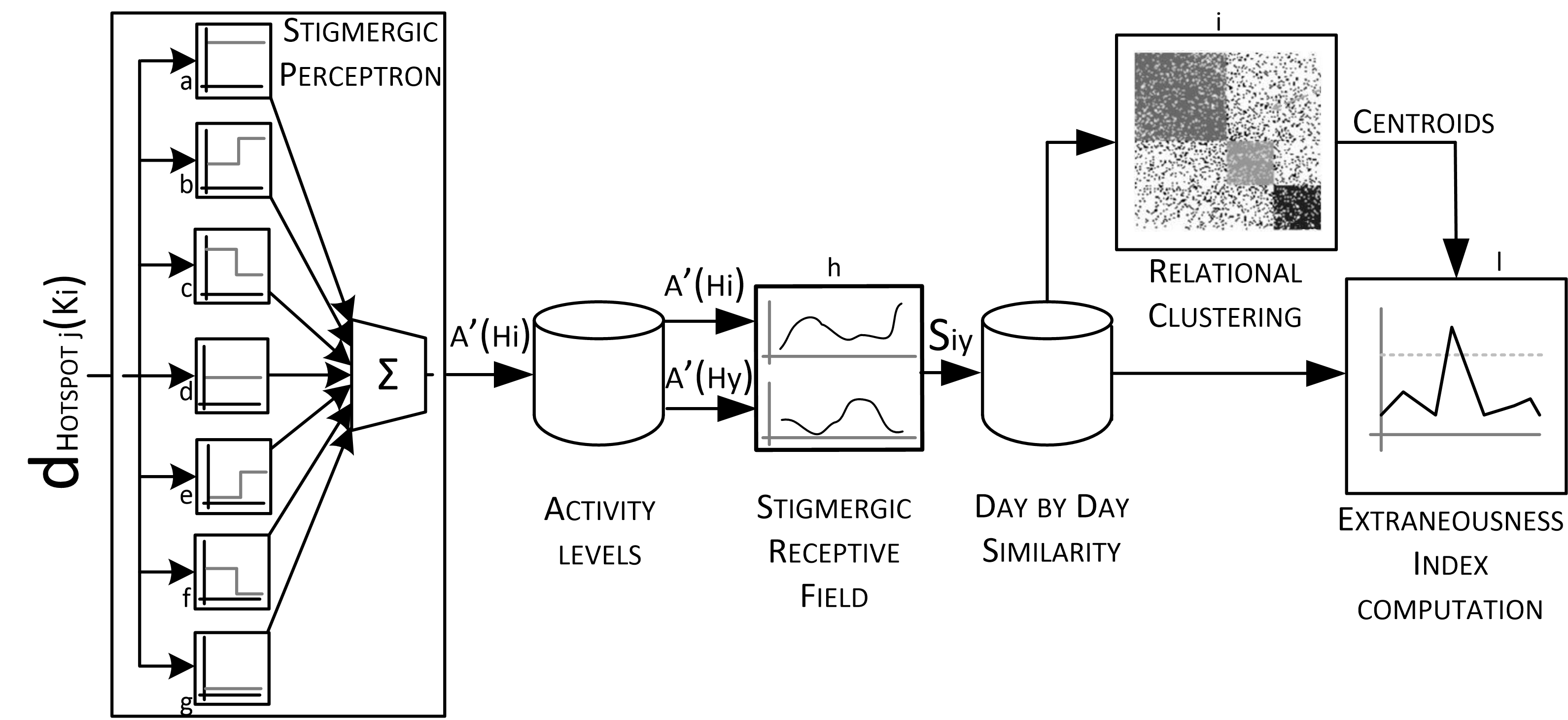}
  \caption{The overall processing of activity samples.}
  \label{img:StigArch}
\end{figure}

Since any real signal is usually similar to more than one archetype, a collection of SRFs, specialized on different archetypes and ordered for increasing activity, is arranged in a connectionist topology to make a Stigmergic Perceptron (Fig. \ref{img:StigArch}). Specifically, adopted archetypes are: 
Asleep (Fig. \ref{img:StigArch}g), i.e. the hotspot at its lowest activity level;
Falling (Fig. \ref{img:StigArch}f), i.e. the flow just before the city activity calms down; 
Awakening (Fig. \ref{img:StigArch}e), i.e. the waking up of urban life after a calm phase;
Flow (Fig.\ref{img:StigArch}d), i.e. the hotspot at its operating capacity, usually exhibited during working hours;
Chill (Fig. \ref{img:StigArch}c), which usually occurs after a rush hour, when people leave work and take taxis to return home; 
Rise (Fig.  \ref{img:StigArch}b), i.e. the hotspot transition to its most intense activity level;
and Rush-Hour (Fig.  \ref{img:StigArch}a), which usually occurs in early morning and late afternoon, when people movement is at its highest rate. 

A perceptron computes a single output from multiple inputs, by forming a linear combination of them. Similarly, the stigmergic perceptron (SP) combines linearly SRFs' outcomes by computing their weighted mean, using the provided similarities $S _{i}$ as weights (Eq. \ref{eq:activity}).

\begin{equation}
\label{eq:activity}
ActivityLevel = \frac{\sum_{i=1}^{N} (S _{i} * i)}{\sum_{i=1}^{N} (S _{i})} 
\end{equation}

The resulting value is called activity level and is defined between zero and N, where N is the number of SRFs. An important aspect concerning hotspot activity level computation is to train each SRF inside a SP in order to prevent multiple activations of SRFs. Let us consider the most sensitive SRF parameter, i.e. the evaporation $\delta$. High evaporation prevents marks aggregation and pattern reinforcement, while low evaporation causes the saturation of the trail. In order to handle this sensitivity, the adaptation of  each SRF inside a SP is twofold: (i) the Global Training phase is aimed to determine an interval for the evaporation rate of each SRF. The interval $[\delta _{min},\delta _{max}]$ is obtained considering the narrowest interval including the fitness values above its 90th percentile, while the intervals for the other parameters can be statically assigned on the basis of application domain constraints; and (ii) the Local Training phase aims to find the optimum values for every module of each single SRF, by using the interval generated in the Global Training phase.
As a result, a proper trained Stigmergic Perceptron provides the characterization of hotspot activity, by transforming a given time series of activity samples in a new time series of activity levels. 
In order to compute the overall similarity between hotspot activity levels gathered in two different days, we employ a further SRF (Fig.  \ref{img:StigArch}h) which uses one activity level time series just like it was an archetype. 
The adaptation in this specific SRF tunes mark width $\epsilon$, trail evaporation $\delta$, and activation thresholds $\alpha_a$ and $\beta_a$. As fitness function, we use the Mean Squared Error (MSE) between computed and ideal similarity over a set of labeled pairs of activity time series (i.e. the training set).

\subsection{Unexpected patterns detection}
Exploiting the mechanism described above, we generate the similarity matrix, that is the collection of similarities obtained by matching with each other the activity level time series of the training set. Provided similarity matrix can be processed by a fuzzy relational clustering technique, grouping days according to their daily activity similarity. Specifically, we employ Fuzzy C-Mean to compute the clusters centroid. The number of clusters corresponds to the number of daily activity behaviors taken into account in the analysis.
Based on these centroids, the membership degrees of further daily activity level time series can be computed. The membership degrees are between 0 (not belonging to the cluster) and 1 (completely belonging to the cluster). By exploiting the membership degrees $u_n$ as a distance, we measure the extraneousness of current activity level with respect to its expected cluster. The Extraneousness Index (EI) is defined as the Manhattan Distance between current daily activity level series $d$ and the centroid of the cluster in which current day is assumed to belong. In Eq. \ref{eq:EI}, the computation case with 3 clusters is shown.
\begin{equation}
\label{eq:EI}
EI(d) = {({\left | u_1(d)-u_1(C_2) \right | + \left | u_2(d)-u_2(C_2) \right | + \left | u_3(d)-u_3(C_3) \right |})} / {2}
\end{equation}
\\

We define as an Unexpected Pattern a day characterized by an activity level whose EI exceeds the maximum EI computed over the training set.

\section{Experimental Studies and Results}
We have analyzed a dataset of taxi traces provided by the Taxi and Limousine Commission of New York City, which contains information about all medallion taxi trips from 2009 to 2016 \cite{dataset}. We focus our investigation on dynamics occurred during 2015 in Manhattan considering that it attracts the most of the taxi trips in New York City. A pre-processing step has been performed to remove missing values and discretize data in spatiotemporal bins defined as a squared area 10-foot- wide with duration of 5 minutes. Then, the min-max normalization is applied. In order to search for hotspots characterizing every possible city routine (i.e. summer and winter ones), the hotspot discovery procedure has been performed comprising data gathered in working days and week-ends of February 2015 and June 2015. 

Since archetypes are assumed to be general, the training set for the SP's global and local phases is generated by using the pure archetype time series as seeds and applying spatial noise and temporal shift. 

In order to validate the SP archetypal behavior detection, a set of time series have been manually labeled and the difference with the actual results of the SP is used to evaluate detection error. Each label corresponds to the expected SP result according to the archetypal behavior visually detected in current time series (i.e. 1 if Asleep, 2 if Falling, and so on). To this purpose, 35 time series (i.e. 5 for each archetype) have been provided to the SP. The obtained MSE is shown in Table \ref{tab:MSE}. By considering the activity level operative range (i.e. [1 7]) and the provided MSE values, the system shows good detection performances, proving the functional effectiveness of the SRF and the SP. 


\begin{table}
\caption{Mean Square Error in Archetypal Behavior Detection via SP.}
\label{tab:MSE}
\begin{center}
\begin{tabular}{c c c c c c c c c}
\hline\rule{0pt}{12pt}
ARCHETYPE & Asleep & Falling & Awakening & Flow & Chill & Rise & Rush-hour & TOT \\
\hline\rule{0pt}{12pt}
MSE &0.215  &0.029	&0.029	&0.028	&0.166	&0.020	&0.143	&0.633 \\
\hline\rule{0pt}{12pt}
\end{tabular}
\end{center}
\end{table}

In the next processing phase, a further SRF is aimed to assess the similarity between daily activity levels. It is provided with a training set obtained by selecting a set of pairs of daily activity levels. In order to supply a clustering process, such SRF is trained to distinguish similar and dissimilar signals, according to the behavioral class of daily activity levels, namely:
(i) Working days (expected to fall between Monday and Tuesday), when crowd movements are mainly caused by working routines; (ii) Entertainment days (expected to fall on Friday and Saturday), in which people tend to spend the night out; (iii) Leisure days (expected to fall on Sunday), which are characterized by limited transportation usage.
Their target similarity is 1 if days belong to the same behavioral class, 0 otherwise. Since the defined classes refer to the cyclical sequence of week days, our ground truth can be provided by the calendar itself. The 10\% of computed daily activity levels have been used to create these pairs (i.e. 1296 pairs overall).

The Fuzzy C-Mean algorithm is used to group days according to their stigmergy-based similarity in order to arrange them among the three provided clusters, namely: Working, Entertainment and Leisure days. Upon this, we exploit the Extraneousness Index in Eq. \ref{eq:EI} to assess unexpected patterns. 

We show results obtained analyzing hotspot D, since it is characterized by multiple usages \cite{zola} allowing the displaying of every activity level behavioral class. Interestingly, this area is also found to be an hotspot by \cite{keler:2016}. 
\begin{figure}
  \centering
  \includegraphics[width=\textwidth]{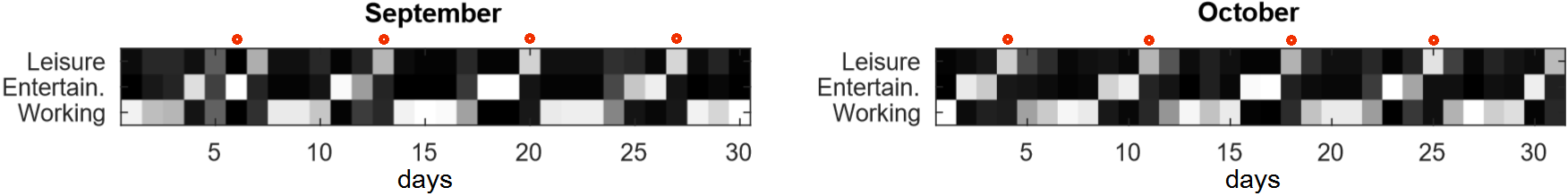}
  \caption{Membership degrees of days in September and October. The whitest, the higher.}
  \label{img:membershipDegree}
\end{figure}

Fig. \ref{img:membershipDegree} shows the computed membership degree for each cluster, obtained with days in September and October. The  whitest the box, the higher the degree. Clearly, the stigmergy-based characterization of hotspot daily activity allows to cluster days according to their behavioral class which corresponds to the arrangement we assumed. Indeed, most of the Sundays (highlighted by a circle in Fig. \ref{img:membershipDegree}) exhibit their highest membership degree with Leisure day cluster. The same happened with days in Entertainment and Working cluster. It is worth noting that provided approach allows the mapping of daily behaviors to emerge from data instead of being explicitly injected into the system.

\begin{figure}
  \centering
  \includegraphics[width=\textwidth]{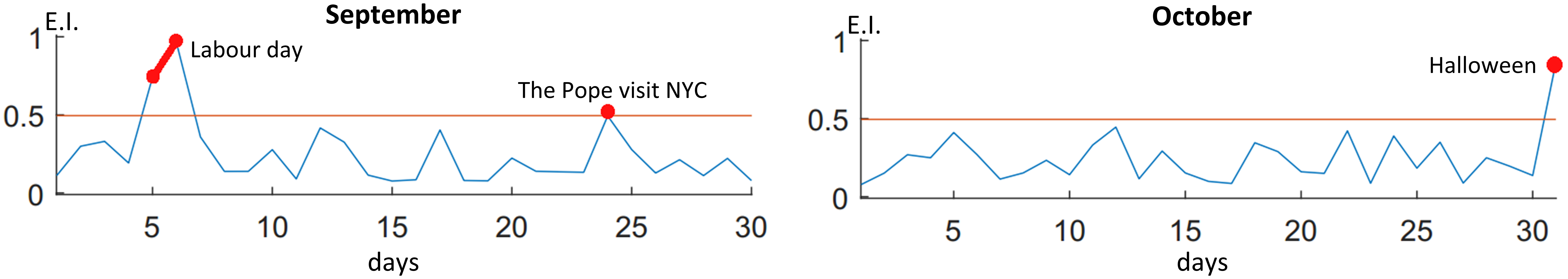}
  \caption{Extraneousness Index computed over days in September and October.}
  \label{img:EI}
\end{figure}

However, some days does not confirm this behavior. Indeed, by comparing their EI with the maximum EI in the training set (red  line in Fig. \ref{img:EI}), they are recognized as an unexpected pattern (red spot in Fig. \ref{img:EI}). 

Table 2 shows the most relevant unexpected patterns detected by analyzing the whole year 2015. Each unexpected pattern date is shown together with their most probable cause, such as an occurred social event. 

\begin{table}
\caption{Most relevant unexpected patterns detected all over 2015.}
\begin{center}
\begin{tabular}{r@{\quad}rl}
\hline
\multicolumn{1}{l}{\rule{0pt}{12pt}
                   EI}&\multicolumn{2}{l}{Date and occurred city event}\\[2pt]
\hline\rule{0pt}{12pt}
0.96                & 06-Sep,       & Labour Day celebration                  	\\
0.94                & 24-May,       & Memorial Day 					   			\\
0.86                & 31-Oct,       & Halloween    					  		 	\\
0.83                & 26-Nov,       & Thanksgiving                          	\\
0.83                & 28-Jun,       & Gay Pride                               	\\
0.82                & 25-Dec,       & Christmas                                	\\
0.81                & 01-Jan,       & New Year's Eve                           	\\
0.80                & 04-Apr,       & Easter (holy Saturday) 					\\
0.79                & 27-Jan,       & Winter Storm Juno \cite{juno}            	\\
0.74                & 05-Sep,       & Labour Day celebrations                  	\\
0.63                & 03-Jul,       & Independence Day                         	\\
0.63                & 31-Dec,       & New Year's Eve                           	\\
0.61                & 15-Mar,       & NYC Half Marathon                      	\\
0.49                & 24-Sep,       & Pope Francis in NYC \cite{pope}  		    \\
\hline
\end{tabular}
\end{center}
\end{table}

EI provides a continuous measure of the magnitude of unexpected patterns, allowing the comparison of their impact on hotspot activity dynamics. As an example, Easter affects the activity in hotspot D much more than the NYC Half Marathon. Indeed, the greatest Easter celebrations in NYC are kept by the St. Patrick Cathedral, which is located in the area corresponding to hotspot D, whereas this area was not directly involved in the NYC Half Marathon 2015. By repeating the analysis in the same date on hotspot C, the computed EI results roughly 60\% higher (i.e. 0.96); indeed the zone corresponding to hotspot C was directly crossed by NYC Half Marathon 2015.

\section{Conclusion}
In this paper, we proposed a novel approach aimed to provide knowledge discovery in the context of human urban mobility data. In contrast with the literature in the field, our approach does not require the in-depth modeling of the dynamics under investigation since it relies on data self-organization provided by employing the principle of stigmergy. Indeed, by using stigmergy, the spatiotemporal density in data has been exploited to identify city hotspots and characterize their dynamics, allowing to generate data-driven prototypes of typical daily activity. By treating them via a clustering technique, we were able to discern expected patterns from unexpected ones, which were found to be usually related to various events. One of the most promising improvements for this investigation can be achieved by cross-checking results obtained via vehicle GPS data with other data sources (e.g. social media or car crash data). Indeed, by employing  a more detailed ground truth, the system can be specialized to model and detect patterns characterized by a timescale shorter than a daily one.


\end{document}